\documentclass[final,3p,12pt,times]{elsarticle}

\usepackage{amssymb}
\usepackage{amsmath}
\usepackage[colorlinks=true,
            linkcolor=blue,      
            citecolor=teal,      
            urlcolor=magenta,    
            pdfborder={0 0 0}    
            ]{hyperref}
\usepackage{booktabs}
\usepackage{xcolor}
\usepackage[most]{tcolorbox}
\usepackage{fancyvrb}
\usepackage{makecell}
\usepackage{pgf-pie}
\usepackage{capt-of}
\usepackage{algorithm}
\usepackage{algpseudocode}
\usepackage[T1]{fontenc}
\usepackage[utf8]{inputenc}
\usepackage{enumitem}
\usepackage{tcolorbox}
\usepackage{alltt}
\usepackage{inconsolata}

\journal{International Journal of Disaster Risk Reduction}

\begin{document}

\begin{frontmatter}

\title{Knowledge-Grounded Agentic Large Language Models for Multi-Hazard Understanding from Reconnaissance Reports}

\author[inst1]{Chenchen Kuai\corref{equal}}
\author[inst1]{Zihao Li\corref{equal}}
\author[inst2]{Braden Rosen}
\author[inst1]{Stephanie Paal}
\author[inst1]{Navid Jafari}
\author[inst1]{Jean-Louis Briaud}
\author[inst1]{Yunlong Zhang}
\author[inst1,inst3]{Youssef M. A. Hashash}
\author[inst1]{Yang Zhou\corref{corauthor}}

\cortext[equal]{These authors contributed equally to this work.}
\cortext[corauthor]{Corresponding author: Yang Zhou \ead{yangzhou295@tamu.edu}.}

\affiliation[inst1]{organization={Zachry Department of Civil \& Environmental Engineering, Texas A\&M University},
            addressline={3136 TAMU}, 
            city={College Station},
            postcode={77843}, 
            state={TX},
            country={USA}}

\affiliation[inst2]{organization={Department of Engineering Technology and Industrial Distribution, Texas A\&M University},
            city={College Station},
            postcode={77843},
            state={TX},
            country={USA}}

\affiliation[inst3]{organization={Department of Civil and Environmental Engineering, University of Illinois Urbana-Champaign},
            city={Urbana},
            postcode={61801},
            state={IL},
            country={USA}}

\begin{abstract}
Post-disaster reconnaissance reports contain critical evidence for understanding multi-hazard interactions, yet their unstructured narratives make systematic knowledge transfer difficult. Large language models (LLMs) offer new potential for analyzing these reports, but often generate unreliable or hallucinated outputs when domain grounding is absent. This study introduces the Mixture-of-Retrieval Agentic RAG (MoRA-RAG), a knowledge-grounded LLM framework that transforms reconnaissance reports into a structured foundation for multi-hazard reasoning. The framework integrates a Mixture-of-Retrieval mechanism that dynamically routes queries across hazard-specific databases while using agentic chunking to preserve contextual coherence during retrieval. It also includes a verification loop that assesses evidence sufficiency, refines queries, and initiates targeted searches when information remains incomplete. We construct HazardRecQA by deriving question–answer pairs from GEER reconnaissance reports, which document 90 global events across seven major hazard types. MoRA-RAG achieves up to 94.5\% accuracy, outperforming zero-shot LLMs by 30\% and state-of-the-art RAG systems by 10\%, while markedly reducing hallucinations across diverse LLM architectures. MoRA-RAG also enables open-weight LLMs to achieve performance comparable to proprietary models. It establishes a new paradigm for transforming post-disaster documentation into actionable, trustworthy intelligence for hazard resilience.

\end{abstract}





\end{frontmatter}

\section{Introduction}
Natural hazards such as earthquakes, hurricanes, floods, and wildfires continue to threaten communities, infrastructure, and economies worldwide~\citep{WuEtAl2025_PostEarthquakeFunctionalRecoveryBridges,verschuur2023multi, zhou2021analyzing}. While each hazard poses distinct risks, many disasters occur in sequence or in combination, amplifying impacts significantly~\citep{ZhangEtAl2023_CascadingHumanRisk}. Earthquakes may destabilize slopes and cause landslides, hurricanes often generate both storm surge and inland flooding, and heavy rainfall can accelerate soil erosion. These complex dynamics make it essential to understand not only individual hazards but also their interactions. A crucial element in building such understanding is the knowledge transfer~\citep{RydstedtNyman2017_SystematicKnowledgeSharing}, the ability to learn lessons and knowledge from past events and apply them to strengthen preparedness, response, and resilience against future hazards.

Knowledge in the hazard domain is preserved in many forms. Operational manuals describe best practices, predictive models simulate hazard scenarios, and sensor networks provide continuous environmental monitoring~\citep{Feng2020SimBasedSeaside, Hart2006_EnvSensorNetworks, PortHouston2020_HurricaneProcedures}. Yet none of these sources fully capture the depth and contextual nuance of post-disaster field evidence. Reconnaissance reports occupy a unique position in this landscape~\citep{GEER2025reportsdb}. Compiled by interdisciplinary teams after major disasters, they document hazard characteristics, infrastructure performance, environmental conditions, and social impacts. These reports integrate measurements, photographs, maps, and expert analyses, often capturing cascading or compound hazards that unfold within a single disaster event. They represent one of the most authoritative sources of hazard knowledge~\citep{Tomac2023}.

However, the very richness that gives these reports their scientific and practical value also makes them challenging to exploit systematically. Reconnaissance documents often span hundreds of pages and encompass a diverse blend of qualitative and quantitative content, including narrative accounts, measurements, maps, tables, and photographic evidence. Although this depth of information offers unparalleled insight into disaster impacts, it is typically presented in highly specialized language filled with technical jargon, abbreviations, and situational references that demand expert interpretation~\citep{Ro2024_PostDisasterDataCuration}. Critical findings, such as the relationships between hazard intensity, structural response, and cascading failures, are frequently fragmented between sections and embedded within unstructured narrative text. The absence of uniform standards among the reporting teams further amplifies this challenge, making it difficult to systematically synthesize, compare, and transfer the knowledge contained in these reports~\citep{lenjani2020towards}.

Recent advances in large language models (LLMs) offer new opportunities to extract, integrate, and reason with this wealth of knowledge. LLMs have demonstrated strong performance on tasks such as summarization, answering questions, and contextual reasoning~\citep{Markowitz2025_KGLLMBench}. Yet, because they rely heavily on broad world knowledge, their output often suffers from hallucinations, producing fluent but unsupported claims. Retrieval-augmented generation (RAG) seeks to address this issue by grounding responses in external documents~\citep{Lewis2020_RAG_KnowledgeIntensiveNLP}. While effective for general-purpose texts, existing RAG frameworks are not well-suited for long and highly technical hazard reports. Once these reports are divided into smaller chunks (i.e., short text segments split for embedding and retrieval) for vector databases, their rich contextual dependencies become fragmented across numerous segments. Retrieving truly relevant evidence from such a vast and heterogeneous hazard corpus remains difficult~\citep{Li2025_LongDocumentRetrievalSurvey}. In addition, most retrieval pipelines rely solely on semantic similarity, tending to retrieve textual information within a single topical dimension ~\citep{Wu2024_ResultDiversification, Karpukhin2020_DensePassageRetrieval}. In hazard domains, this corresponds to retrieving information from one hazard type while neglecting multiple or cascading hazards. Consequently, the retrieved information can be incomplete or misleading, and the LLMs built upon it often fail to produce trustworthy, well-grounded responses. Moreover, conventional RAG systems lack mechanisms to assess whether the retrieved evidence is sufficient or to recognize when a reliable answer cannot be provided~\citep{Joren2024_SufficientContext_RAG}.

These limitations reveal a fundamental gap between current retrieval architectures and the demands of multi-hazard reasoning. Addressing this gap requires a framework capable of both precision retrieval and iterative validation across diverse hazard domains. To this end, this paper presents the first systematic effort to adapt RAG for transforming hazard reports, such as reconnaissance documents, into structured and reliable world knowledge that supports multi-hazard understanding and reasoning with LLMs. A dedicated dataset, \textit{HazardRecQA}, is developed from reconnaissance reports to benchmark the quality of knowledge transfer. Building upon this foundation, we propose the \textit{Mixture-of-Retrieval Agentic RAG} (MoRA-RAG) framework, which overcomes the limitations of existing approaches. MoRA-RAG incorporates a \textit{Mixture-of-Retrieval} (MoR) mechanism that dynamically selects relevant hazard databases through a routing strategy, enhancing retrieval precision and highlighting cross-hazard relationships. Its agentic architecture actively verifies the sufficiency and consistency of retrieved evidence, performs targeted online searches to fill knowledge gaps, and synthesizes external information to strengthen reasoning robustness. By combining these capabilities, MoRA-RAG establishes a new paradigm for trustworthy and context-aware knowledge transfer in the hazard domain, enabling LLMs to reason across interconnected disaster processes and support more informed decision-making.

\section{Related Literature}

\subsection{Knowledge transfer in natural hazards}

Knowledge transfer from past events helps understand how hazards evolve, interact, and impact communities as new events unfold ~\citep{zhou2021analyzing,kuai2025us}. By drawing on prior experiences, researchers and practitioners are able to identify patterns of vulnerability, recognize emerging risks, and anticipate cascading effects across interconnected systems ~\citep{brunner2024understanding,spiekermann2015disaster, quitana2020resilience}. Forecasting and predictive modeling leverage data from historical events to identify recurring patterns and prepare in advance of upcoming hazards, forming the foundation of modern early warning systems ~\citep{paudel2025predicting,crawford2018risk}. Similarly, resilience analyses based on past disasters reveal how infrastructure, logistics, and social systems respond and recover under disruption, offering insights into system fragility and redundancy ~\citep{purwar2024improving}. All together, knowledge transfer remains a cornerstone of hazard impact analysis and an indispensable component of disaster preparedness and response.


However, much of this knowledge transfer still relies primarily on structured data (e.g., tabular records and quantitative indicators). At the same time, a substantial portion of valuable information contained in heterogeneous and unstructured sources remains underutilized ~\citep{nuwara2025unlocking,li2025leveraging,zhou2025automated}. Reconnaissance reports, field observations, and expert narratives often provide rich contextual details about on-the-ground conditions, response actions, and cascading impacts that are difficult to capture through structured datasets alone~\citep{yesiller2023disaster}. Yet, processing such heterogeneous data is inherently challenging. Only a limited number of studies have attempted to bridge these unstructured sources using advanced natural language processing (NLP) and related analytical techniques. For instance, Ma et al.~\citep{ma2022text} adopted text mining and extracted key information from dense geological hazard documents, while He et al.~\citep{he2024impact} integrated past textual information into models of social response during disasters through event-embedding approaches. While effective, existing efforts remain largely task-specific and domain-bounded. A recent review paper calls for scalable disaster information management and highlights the importance of transforming knowledge from these rich yet heterogeneous sources; however, still no practical framework has been developed ~\citep{xu2025llm_disaster}.



\subsection{Evolution of LLMs and context engineering}

The rapid development of LLMs has drastically reshaped NLP capabilities and opened new opportunities to bridge existing gaps in hazard knowledge transfer from unstructured reports. Transformer-based foundation models such as ChatGPT, Claude, Gemini, and LLaMA exhibit strong general abilities in comprehension, summarization, and question answering ~\citep{achiam2023gpt4, team2023gemini, touvron2023llama}. Models such as DeepSeek have further enhanced reasoning performance through reinforcement learning-based optimization~\citep{guo2025deepseekr1,gibney2025secrets}. However, these foundation models are primarily trained on broad, general-purpose text and remain insufficiently grounded in domain knowledge (e.g. hazard knowledge)~\citep{hung2023limitations}. As a result, they often generate hallucinated or incomplete responses when applied to domain-specific tasks. To mitigate this issue, context engineering has emerged as a promising solution. Context engineering refers to designing an information-rich environment, such as incorporating domain-specific documents, structured metadata, or tailored prompts, so that LLMs are better grounded when generating answers~\citep{mei2025survey_context_engineering, zheng2025survey_knowledge_augmented}.


In practice, the context engineering is often realized through RAGs. RAG augments the context of LLMs with retrieved text passages, leading to large improvements in knowledge-intensive tasks~\citep{asai2023selfrag}. Recent studies have also applied RAG for hazard and climate resilience, such as ChatClimate~\citep{vaghefi2023chatclimate}, which enables question answering and reasoning about climate-related information, and other systems that assist analysis and decision-making during natural hazards and extreme weather events ~\citep{juhasz2024responsible}. While promising, RAG faces critical challenges: when irrelevant or low-quality evidence is retrieved, model performance can degrade rather than improve ~\citep{merth2024superposition,ouyang2025hoh}. This “garbage-in, garbage-out” issue underscores persistent limitations in retrieval accuracy and evidence reliability. Recent models such as C-RAG~\citep{yan2024corrective} and MAIN-RAG~\citep{chang2024mainrag} introduce correction and validation mechanisms to refine retrieved knowledge, but they mainly focus on re-actively correcting retrieval errors. Mechanisms for sufficiency and reliability of the information to answer the query remain unexplored.

\section{QA Dataset Construction}


HazardRecQA is developed to evaluate an LLMs’ ability to transfer knowledge from post-disaster reconnaissance documents through question–answer (QA) pairs. These QA pairs provide the foundation for assessing model performance (e.g., LLM and RAG)~\citep{wang2023evaluating}, where the answering accuracy indicates how effectively a model captures and applies hazard knowledge from the reports. The ground-truth hazard knowledge is obtained from post-disaster reconnaissance reports (i.e., GEER Reports~\citep{GEER2025reportsdb}) and organized into QA form for systematic evaluation.

\begin{figure}[ht]
    \centering
    \includegraphics[width=0.80\columnwidth]{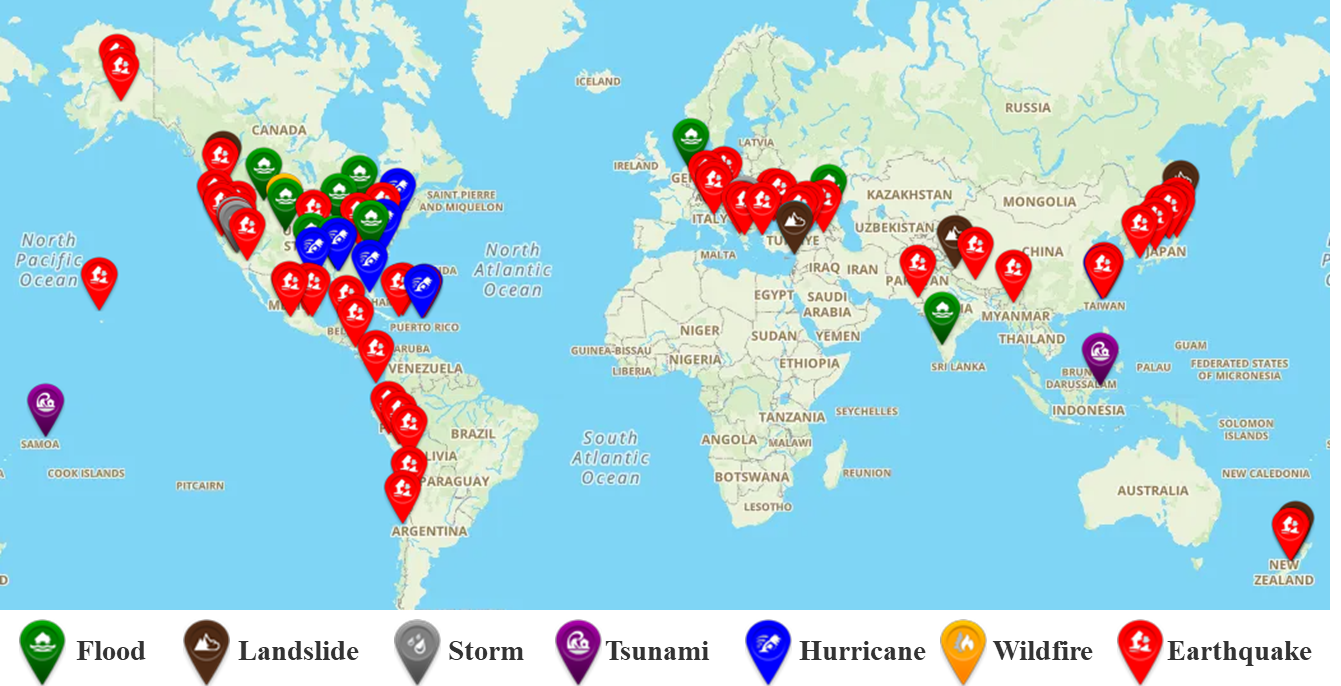}
    \caption{Coverage and event distribution of GEER reconnaissance reports~\citep{GEER2025reportsdb}.}
    \label{fig:GEER}
\end{figure}

GEER reconnaissance reports provide comprehensive post-disaster documentation, including site observations, photographs, instrumentation data, geospatial records, and analyses of failure mechanisms. These reports span seven categories of natural hazards\footnote{Hurricanes and typhoons are considered the same hazard type, as they are regional names for the same phenomenon. In this study, both are collectively referred to as hurricanes.} observed worldwide from 1989 to 2025 (Figure~\ref{fig:GEER}). As the world’s largest and most comprehensive open-access collection of post-disaster field reconnaissance, the GEER database encompasses 58 earthquakes, 9 floods, 2 tsunamis, 1 wildfire, 4 storms, 5 landslides, and 11 hurricanes, totaling 90 hazard events. Building on the knowledge preserved in these reports, the questions in HazardRecQA are designed to evaluate models across four categories of hazard-related knowledge: (1) \textit{Hazard Characteristics}: describing the event itself, including what happened, when and where it occurred, its magnitude, setting, or physical triggers; (2) \textit{Analysis Approach}: focusing on how the event was examined through data collection, instruments, surveys, models, or analytical methods; (3) \textit{Impacts and Damage}: addressing the consequences or losses caused by the event, such as structural failures, economic damage, or human impacts; and (4) \textit{Response and Recovery}: covering the actions taken after the event, including emergency response, evacuation, repair, or long-term recovery.

The QA generation process employs GPT-5 nano to automatically construct factual QA pairs from reconnaissance report paragraphs, following recent advances showing that LLMs can comprehend complex domain text and autonomously generate coherent, fact-based questions~\citep{ehsan2025automatic,Li2024PFQS}. Each paragraph from the reconnaissance reports undergoes a data cleansing step to remove non-informative components (e.g., tables of contents, acknowledgments, references) and retain semantically rich content for question construction. GPT-5 is then guided by a structured instruction with strict output constraints to generate verifiable QA pairs in either True/False or Multiple-Choice (A–D) formats. These formats are adopted because they provide clear factual grounding and allow objective evaluation through accuracy-based metrics~\citep{chan2023case}. Each QA item is further classified into the four question categories introduced above. The generated outputs undergo post-validation through schema checking to ensure consistency with the predefined QA formats.

The HazardRecQA dataset contains a total of 5,776 QA pairs generated from post-disaster reconnaissance reports. As shown in Table \ref{tab:qtype-category} and Figure \ref{fig:hazard-pie}, each QA instance is categorized along two dimensions: (1) task category, including Hazard Characteristics, Response and Recovery, Analysis Approach, and Impacts and Damage, and (2) hazard type, such as floods, landslides, storms, tsunamis, hurricanes, wildfires, and earthquakes. The dataset comprises both True/False and Multiple-Choice formats, enabling diverse factual tasks and straightforward evaluation through answer accuracy. The complete QA construction prompts are presented in \ref{first: QA}.


\begin{figure*}[t]
  \centering

  \begin{minipage}[b]{0.48\textwidth}
    \centering
    \footnotesize
    \setlength{\tabcolsep}{6pt}
    \begin{tabular}{@{}lccc@{}}
      \toprule
      \textbf{Task Category} & \textbf{TF} & \textbf{MC} & \textbf{ALL} \\
      \midrule
      Hazard Characteristics & 1652 & 692 & 2344 \\
      Response and Recovery  & 123  & 83  & 206  \\
      Analysis Approach       & 666  & 844 & 1510 \\
      Impacts and Damage      & 1175 & 541 & 1716 \\
      \midrule
      \textbf{Total}          & \textbf{3616} & \textbf{2160} & \textbf{5776} \\
      \bottomrule
    \end{tabular}
    \vspace{0.4em}
    \captionof{table}{Question type distribution across task categories, TF (True/False) and MC(Multiple-Choice).}
    \label{tab:qtype-category}
  \end{minipage}
  \hfill
  \begin{minipage}[b]{0.48\textwidth}
    \centering
    \footnotesize
    \begin{tikzpicture}
      \pie[
        sum=auto,
        radius=1.4,
        text=legend,              
        before number=\phantom,   
        after number=,            
        color={cyan!40, orange!45, lime!40, violet!40, pink!45, gray!40, teal!35}
      ]{
        9.1/Flood (658),
        3.6/Landslide (263),
        4.8/Storm (348),
        1.0/Tsunami (70),
        12.4/Hurricane (891),
        1.3/Wildfire (92),
        67.8/Earthquake (3454)
        }
    \end{tikzpicture}
    \vspace{0.4em}
    \captionof{figure}{Proportional distribution of total valid QA across hazard types (total QA).}
    \label{fig:hazard-pie}
  \end{minipage}

\end{figure*}

\section{Methodology}




This section will first introduce the framework of RAG, explaining how external documents are processed into vector databases and how the retrieval and generation modules function to produce grounded answers. We then extend this framework through the proposed Mixture-of-Retriever Agentic RAG (MoRA-RAG), which enhances both database construction and retrieval precision. Furthermore, an agentic LLM framework is incorporated to validate, reflect on, and refine the generated responses for improved reliability.

\subsection{Retriever-Argumented Generation}
\begin{figure}[ht]
    \centering
    \includegraphics[width=0.85\columnwidth]{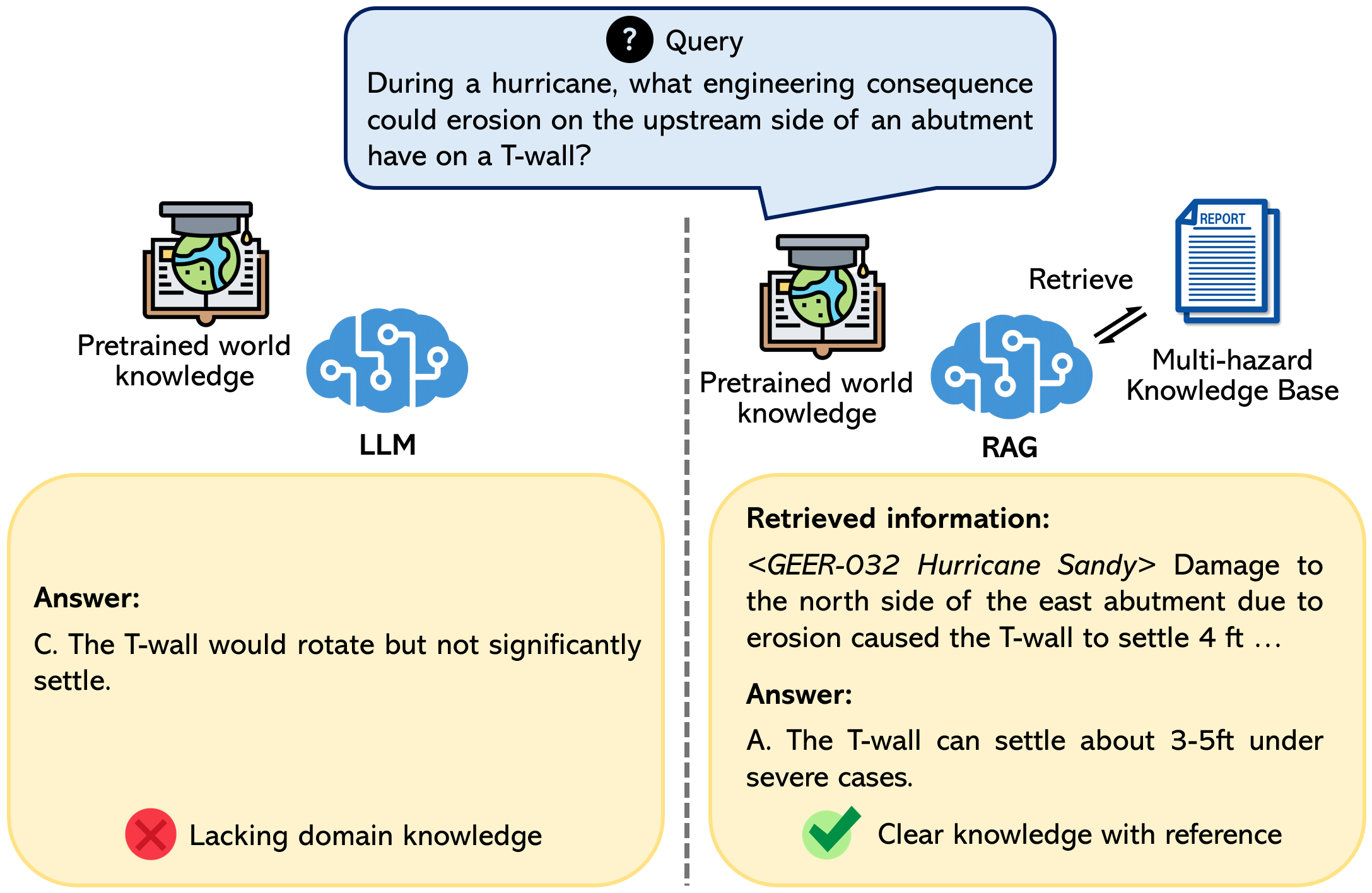}
    \caption{RAG systems explained, it is able to extract and transfer hazard knowledge from external databases.}
    \label{fig:RAG_demo}
\end{figure}

The RAG framework operates in two main stages: (1) constructing a knowledge database from external documents, and (2) performing retrieval and answer generation based on a given query. As illustrated in Figure~\ref{fig:RAG_demo}, consider an example question asking about the engineering consequence of erosion on the upstream side of an abutment during a hurricane. When relying solely on its general world knowledge, the LLM produces a generic response that lacks domain-specific evidence. In contrast, RAG retrieves relevant information from multiple reconnaissance reports and identifies a specific segment in the GEER-032 Hurricane Sandy Report, which documents abutment damage and T-wall settlement. By grounding the generation in such retrieved evidence, RAG enables the model to effectively transfer knowledge from the documents.

\subsubsection{Database chunking}

In RAG, the first step is to construct a vector database suitable for retrieval. The original documents (e.g., reconnaissance reports) are often lengthy and cannot be directly fed into a model’s context window. Therefore, the documents are required to be segmented into smaller text units; this step is known as chunking. These chunks are later converted into vector representations and are jointly preserved in the vector database. 

We denote the chunking process the documents \( \mathcal{D} \) as:

\begin{equation}
\{d_1, d_2, \dots, d_n\} = \varphi(\mathcal{D})
\end{equation}

where \( \{d_1, d_2, \dots, d_n\} \) represents the set of chunks produced by the chunker \( \varphi \) from document \( \mathcal{D} \). Common strategies include fixed-token and paragraph-based chunking~\citep{lewis2020rag,bhat2025rethinking}. While widely used, these methods can break the logical structure of the text, causing key contextual relationships, such as cause-and-effect connections or hazard impact descriptions, to be split across segments~\citep{jimeno2024financial}. This fragmentation often weakens retrieval relevance and reduces the completeness of the retrieved information. 

Each chunked text is then converted into a numerical vector representation using an embedding function \( f(\cdot) \):
\begin{subequations}
\begin{align}
e &= f(d) \\
f(\cdot): \mathcal{X} &\rightarrow \mathbb{R}^{m}
\end{align}
\end{subequations}
where \( \mathcal{X} \) denotes the text space and \( \mathbb{R}^{m} \) represents the \(m\)-dimensional embedding space determined by the language embedding model.
\texttt{text-embedding-3-small}~\citep{openai2025embedding}, an embedding model from OpenAI adopted in this study, produces 1,536-dimensional embeddings.

Finally, each chunk and its corresponding embedding are stored in the vector database \( \hat{\mathcal{D}} \):

\begin{equation}
\hat{\mathcal{D}} = \{(d_1,e_1), (d_2,e_2), \dots, (d_n,e_n)\}
\end{equation}

This database serves as the external knowledge source that supports later retrieval and generation within the RAG framework.

\subsubsection{Retriever \& Generation}

The retrieval component in RAG is responsible for locating the most relevant information from the external vector database \( \hat{\mathcal{D}} \), while the generation component integrates this retrieved knowledge into the LLM’s context to produce grounded and reliable answers. Formally, the RAG framework, denoted as \( \mathcal{M} \), can be expressed as:

\begin{equation}
\mathcal{M} = \big(\mathcal{G}, \mathcal{R}\big), \quad
\mathcal{M}(q \mid \hat{\mathcal{D}}) = \mathcal{G}\big(q, \mathcal{R}(\hat{\mathcal{D}} \mid q)\big)
\end{equation}

where \( \mathcal{G} \) is the generation module (i.e., the backbone LLM that generates the response) and \( \mathcal{R} \) is the retrieval module. Given a user query \( q \) and an external knowledge base \( \hat{\mathcal{D}} \), the retrieval module identifies and returns the most relevant evidence \( \mathcal{R}(\hat{\mathcal{D}} \mid q) \), which is then used by \( \mathcal{G} \) to generate the final answer.

The retriever first identifies the most relevant evidence from the database with respect to the query \( q \) through a two-stage process: a coarse-grained retrieval followed by a fine-grained reranking. These consist of a fast, 'coarse-grained' search to find potential matches, followed by a slower, 'fine-grained' reranking to find the best matches. The coarse-grained stage is performed by a bi-encoder, which encodes the query and each document independently and measures their similarity efficiently, though at lower precision. The fine-grained stage then uses a cross-encoder to refine the ranking of retrieved candidates with higher accuracy but greater computational cost. The bi-encoder and cross-encoder will be introduced in detail. The retriever therefore, balances the performance and computational cost. It first selects the top-\( L \) candidate chunks using the bi-encoder, where \( L \) controls the number of chunks passed to the cross-encoder for reranking. The value \( L = 50 \) is adopted following prior work~\citep{lewis2020rag}. The cross-encoder then reranks these \( L \) candidates and selects the top-\( K \) most relevant chunks, with \( K = 5 \) adopted~\citep{yu2024rankrag}.

Formally, the bi-encoder computes the cosine similarity between the embeddings of the query and each chunk, and selects top--\( L \) chunks with highest similarity score:

\begin{subequations}
\begin{align}
\{d_1, d_2, \dots, d_L\} &= \operatorname*{arg\,max}_{d \in \hat{\mathcal{D}}}^{L} \, \mathcal{L}_b\!\left(d,q\right), \\
\mathcal{L}_b \big(d,q\big) &= \frac{e \cdot f(q)}{\| e \| \, \| f(q) \|},
\end{align}
\end{subequations}

where \( f(q) \) and \( e \in \hat{\mathcal{D}} \) are the embedding vectors of the query and chunk, respectively. The cosine similarity function \( \mathcal{L}_b(\cdot) \) measures the normalized inner product between the two embeddings, with higher values indicating stronger semantic relevance.

In the fine-grained, reranking stage, the cross-encoder takes both the query and each candidate chunk as a single combined input, allowing direct interaction between the two. This model jointly encodes the pair into a contextual representation \( g(q, d) \), and the relevance score is computed as:

\begin{subequations}
\begin{align}
\{d_1, d_2, \dots, d_K\} &= \operatorname*{arg\,max}_{d \in \{d_1, d_2, \dots, d_l\}}^{K} \, \mathcal{L}_c\!\left(d,q\right), \\
\mathcal{L}_c \big(d,q\big) &= \sigma \big(\mathbf{w}^\top g(q,d)\big),
\end{align}
\end{subequations}

where \( \sigma(\cdot) \) denotes the sigmoid activation and \( \mathbf{w} \) is a learnable scoring vector. The resulting top-\( k \) chunks represent the most contextually relevant evidence retrieved for the query.

The final retrieval output concatenates these selected chunks to form the context used for generation:

\begin{equation}
\mathcal{R}(\hat{\mathcal{D}} \mid q) = d_1 \oplus d_2 \oplus \cdots \oplus d_k
\end{equation}

This retrieved evidence \( \mathcal{R}(\hat{\mathcal{D}} \mid q) \) is then added to the original query, enabling the generation module \( \mathcal{G} \) to produce the evidence-grounded answer.

\subsection{Mixture-of-Retriever Agentic RAG}

\begin{figure}[ht]
    \centering
    \includegraphics[width=0.98\columnwidth]{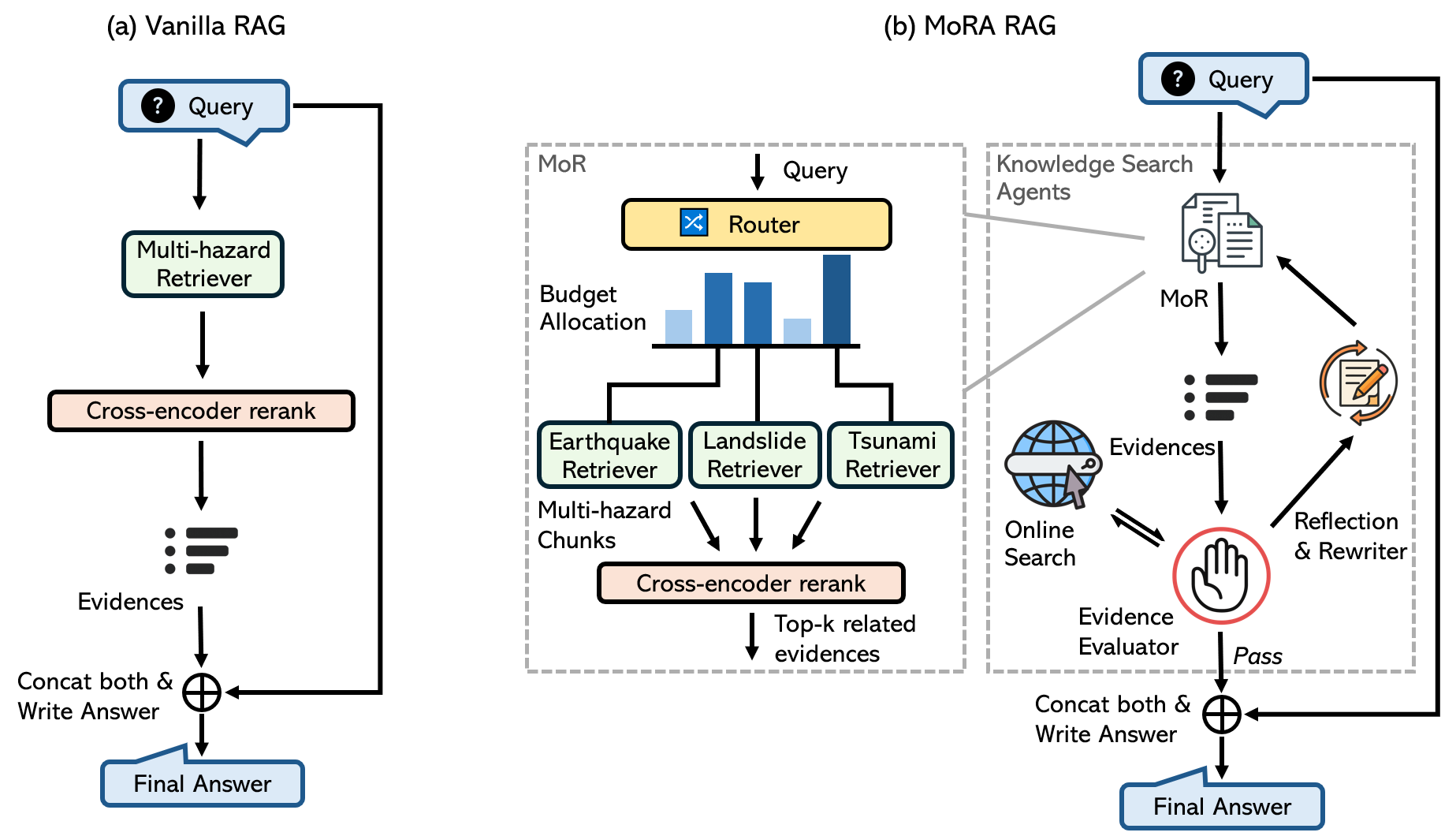}
    \caption{Architecture of the proposed MoRA-RAG framework compared with Vanilla RAG.}
    \label{fig:your_label}
\end{figure}

We introduce the Mixture-of-Retriever Agentic RAG (MoRA-RAG) framework (Figure~\ref{fig:your_label}), which extends the standard RAG by integrating two key enhancements. First, it incorporates an improved chunking strategy and a mixture-of-retriever mechanism to increase retrieval precision. Second, it introduces an agentic LLM framework designed to address the validation gap in traditional RAG systems. This agentic framework enables the model to validate, reflect upon, and refine its retrieved evidence through coordinated specialized agents. The detailed design and function of each component are presented in the following sections.

\subsubsection{Agentic Chunk}
In MoRA-RAG, the chunking process is improved by incorporating LLM agents as the document chunker \( \varphi \), allowing the segmentation of reconnaissance reports into knowledge-preserving units. Instead of relying on fixed-token or paragraph-based segmentation, which may fragment the logical flow of hazard observations, the agentic chunker ensures that each chunk maintains contextual completeness and factual clarity.

The process begins with the extraction of propositions, where the LLM identifies concise factual statements from long paragraphs. Each proposition is rewritten into a standalone and explicit sentence that includes both subject and predicate, for example, transforming “Observed liquefaction along quay walls” into “Investigators observed liquefaction along quay walls.” Next, the LLM groups related propositions into compact and thematically consistent chunks. Each chunk typically combines statements describing hazard conditions, observed impacts, and contextual factors, followed by a short LLM-generated summary capturing the key idea. To maintain focus and internal consistency, each chunk contains fewer than ten propositions, avoiding mixed or overly long segments. This structure reflects how domain experts summarize multiple related observations in post-disaster reports.

The resulting chunks are further paired with their embedding and produce the vector database, denoted as \( \hat{\mathcal{D}} \). This serves as the enhanced knowledge database for retrieval and generation in MoRA-RAG. Implementation details and comparisons with basic chunking methods are detailed in~\ref{sec:chunk}.

\subsubsection{Mixture of Retriever}
The MoR module enhances the retrieval process by enabling hazard-aware routing and adaptive evidence selection from the database. Instead of retrieving from a single unified knowledge base, the query is first analyzed by an router agent that estimates the relevance distribution over multi-hazard domains. This design allows the framework to allocate retrieval budgets, focusing on the hazards most likely related to the query, while maintaining coverage over potentially interacting hazard types.


Formally, let \( \mathcal{H} \) denote the complete set of hazard categories (e.g., earthquake, flood, storm, landslide, wildfire, tsunami). The router agent estimates a probability vector \( \mathbf{p} = \mathcal{T}(q) \), where each component \( p_h \) represents the likelihood that the query \( q \) is associated with hazard \( h \), and all probabilities sum to 1. A threshold \( \tau \) is then applied to filter out unrelated hazards, yielding the subset of relevant hazards \( \mathcal{S} = \{\, h \mid p_h \ge \tau,\, h \in \mathcal{H}\,\} \). In this study, a threshold of \( \tau = 0.2 \) is adopted to strike the balance of potential information loss and computational demand. Given the fixed bi-encoder retrieval budget \( L \) (same as introduced in RAG), the router proportionally assigns the chunk budget \( l_h \) to each selected hazard category \( h \) according to its probability:

\begin{equation}
l_h = \frac{p_h}{\sum_{j \in \mathcal{S}} p_j} \, L, \qquad h \in \mathcal{S}.
\end{equation}


With the router, the retrieval budget is dynamically distributed across hazards according to their relevance scores. Each hazard-specific database \( \hat{\mathcal{D}}_h \) then performs retrieval independently under its allocated quota.

During the coarse-grained retrieval stage, each hazard-specific retriever selects the top-\( l_h \) chunks based on the cosine similarity between the query and the chunk: 

\begin{equation}
\{ d_{h1}, d_{h2}, \dots, d_{hl_h} \}
  = \operatorname*{arg\,max}_{d \in \hat{\mathcal{D}}_h}^{l_h}
    \mathcal{L}_b \big(d,q\big),
\end{equation}

The retrieved chunks from all relevant hazards are then aggregated into the candidate chunks set \( \mathcal{C} = \bigcup_{h \in \mathcal{S}} \{ d_{h1}, \dots, d_{hl_h} \} \), representing the multi-hazard evidence set for refinement.  

A fine-grained reranking stage uses a cross-encoder that takes the query and a candidate chunk as a single combined input and encodes them together. The final retrieval output selects the top-\( k \) chunks across the multi-hazard candidate chunks: 

\begin{equation}
\mathcal{R}(\mathcal{D} \mid q)
  = \operatorname*{arg\,max}_{d \in \mathcal{C}}
    \mathcal{L}_{c} \big(d,q\big).
\end{equation}

\subsubsection{Agentic LLM Structure}

The agentic LLM structure functions as the information validation and enhancement layer beyond the Mixture-of-Retrieval (MoR) module. Its primary goal is to verify the quality of retrieved evidence and continuously refine it through iterative loops until trustworthy information is obtained.

This framework consists of five cooperative agents that collectively manage evidence retrieval, validation, reflection, and external data search. The overall workflow is illustrated in Algorithm \ref{alg:mora_rag}, where the left column outlines the procedural pipeline and the right column highlights the agents operating at each stage. In each iteration, the MoR retriever first gathers hazard-aware evidence from the relevant databases. The Evidence Evaluator then examines whether this evidence is sufficient to answer the user’s query. If not, the Online Search Agent supplements the evidence with external information sources. When both in-domain and external evidence remain inadequate, the Reflection and Question Rewriter agent revises the query, prompting another retrieval cycle. The process iterates until reliable evidence is obtained or the iteration limit is reached. We set the limit to 5 to balance accuracy and computational cost. Finally, the Answer Writer synthesizes the validated information into a grounded response.

\begin{algorithm}[h]
\caption{\textbf{Agentic MoRA-RAG Inference}}
\label{alg:mora_rag}
\begin{algorithmic}[1]
\Require Generation module \( \mathcal{G} \); MoR retriever module \( \mathcal{R} \);
Online search agent \(O\); Evidence evaluator \(E\);
Reflection \& Question rewriter \(Q\)
\State \textbf{Input:} User query \(q\), Multi-hazard databases \(\{\hat{\mathcal{D}}_h\}\)
\State \textbf{Output:} Grounded answer \(y\)
\For{iteration \(t = 1\) to \(5\)}
  \State MoR module retrieves hazard-aware evidence $C = \mathcal{R}(q,\{\hat{\mathcal{D}}_h\})$ \hfill $\triangleright$ \textcolor{blue}{MoR Router}
  \State Evaluator checks Sufficiency of $C$ to answer $q$ \hfill $\triangleright$ \textcolor{blue}{Evidence Evaluator}
  \If{Sufficiency == Yes}
      \State Evidence $E = C$; \textbf{break}
  \ElsIf{Sufficiency == No}
      \State Online Search Agent retrieves external evidence $C_o = O(q)$ \hfill $\triangleright$ \textcolor{blue}{Online Search}
      \State Evaluator checks Sufficiency of $C_o$ to answer $q$ \hfill $\triangleright$ \textcolor{blue}{Evidence Evaluator}
      \If{Sufficiency == Yes}
          \State Evidence $E = C_o$; \textbf{break}
      \Else
          \State Reflect and rewrite query $q = Q(q,C)$ \hfill $\triangleright$ \textcolor{blue}{Reflection \& Question Rewriter}
      \EndIf
  \EndIf
\EndFor
\If{Evidence $E$ exists}
  \State $y = \mathcal{G}(q, E)$ \hfill $\triangleright$ \textcolor{blue}{Answer Writer}
\Else
  \State $y = \mathcal{G}(q)$ \hfill $\triangleright$ \textcolor{blue}{Answer Writer (Fallback)}
\EndIf
\State \textbf{Return:} final answer $y$
\end{algorithmic}
\end{algorithm}

The agents serve distinct yet complementary purposes: (1) \textit{MoR Router Agent} routes hazard-related queries by estimating a probability distribution over all hazard categories, determining which knowledge bases or retrievers should be prioritized; (2) \textit{Online Search Agent} performs lightweight web-based retrieval using external search APIs (i.e., qdrant-client~\citep{qdrant2025}) to supplement online knowledge when local retrieval is insufficient; (3) \textit{Evidence Evaluator Agent} examines the sufficiency and relevance of retrieved evidence, deciding whether it is sufficient to answer the question (output ``1'') or if more information is required (output ``0''); (4) \textit{Reflection \& Question Rewriter Agent} reviews the retrieved content and reformulates ambiguous or ineffective queries to improve retrieval precision through iterative reflection and rewriting; and (5) \textit{Answer Writer Agent} synthesizes validated evidence into final grounded responses, ensuring factual accuracy and coherence across multi-hazard contexts. To ensure reproducibility, detailed prompt designs for all agents are provided in \ref{third:agent}.

\section{Experiments}

This section outlines the experimental design used to evaluate the proposed framework. We first test model performance in the zero-shot setting, where LLMs rely solely on pretrained knowledge without external input. This serves as a baseline for understanding their inherent reasoning capacity in multi-hazard contexts. We then introduce the RAG setting to examine how adding report-based knowledge improves factual grounding and contextual understanding. While this enhances overall accuracy, limitations remain when retrieved evidence is incomplete or unrelated to the query. Building on these observations, we evaluate the proposed MoRA-RAG with state-of-the-art RAG frameworks. These benchmarks enable a direct comparison with the proposed approach, allowing us to assess whether its mixture-of-retriever design and agentic verification loop lead to measurable improvements in retrieval quality and answer accuracy.

An ablation study is also conducted to examine the contribution of key components within the proposed framework. 
The analysis is structured around two research questions (RQs):
\begin{itemize}
    \item \textbf{RQ1:} How does the agentic structure contribute to more effective hazard understanding, and what is the role of each module in improving knowledge extraction and generation?
    \item \textbf{RQ2:} How do different chunking strategies affect the reliability and efficiency of knowledge retrieval across multi-hazard databases?
\end{itemize}

\subsection{Baseline Models and Metric}
The evaluation includes two settings: zero-shot and RAG-based. In the zero-shot setting, LLMs generate answers using only their pretrained knowledge without external input. We tested two open-weight models (Gemma series and GPT-oss) and three proprietary models (Gemini-2.5-Flash ~\citep{team2023gemini}, GPT-5-Nano  ~\citep{achiam2023gpt}, and Claude-sonnet-4~\citep{anthropic2024claude}). The Gemma series ~\citep{mesnard2024gemma}, ranging from 1B to 27B parameters (1B, 4B, 12B, and 27B), was selected to examine how model complexity influences performance. In the RAG-based setting, three retrieval frameworks were included for comparison: Vanilla RAG, CRAG (Corrective Retrieval-Augmented Generation~\citep{yan2024corrective}), and MAIN-RAG (Multi-Agent Filtering Retrieval-Augmented Generation~\citep{chang2024mainrag}). These models serve as benchmarks for evaluating the performance of the proposed MoRA-RAG. All experiments were conducted using an NVIDIA A100 GPU for open-weight LLMs and via official APIs for proprietary models.

All questions in the \textit{HazardRecQA} dataset are true/false or multiple-choice. Model performance is evaluated using accuracy~\citep{kuai2025cyportqa}, defined as

\begin{equation}
\text{Accuracy} = \frac{1}{N} \sum_{i=1}^{N} \mathbb{I}(\hat{y}_i = y_i)
\end{equation}

where \( N \) is the total number of questions, \( y_i \) is the ground-truth answer, \( \hat{y}_i \) is the LLMs’ answer, and \( \mathbb{I}(\cdot) \) equals 1 if the prediction is correct and 0 otherwise. 

\subsection{Model performances}

\begin{figure}[ht]
   \centering
   \includegraphics[width=0.98\columnwidth]{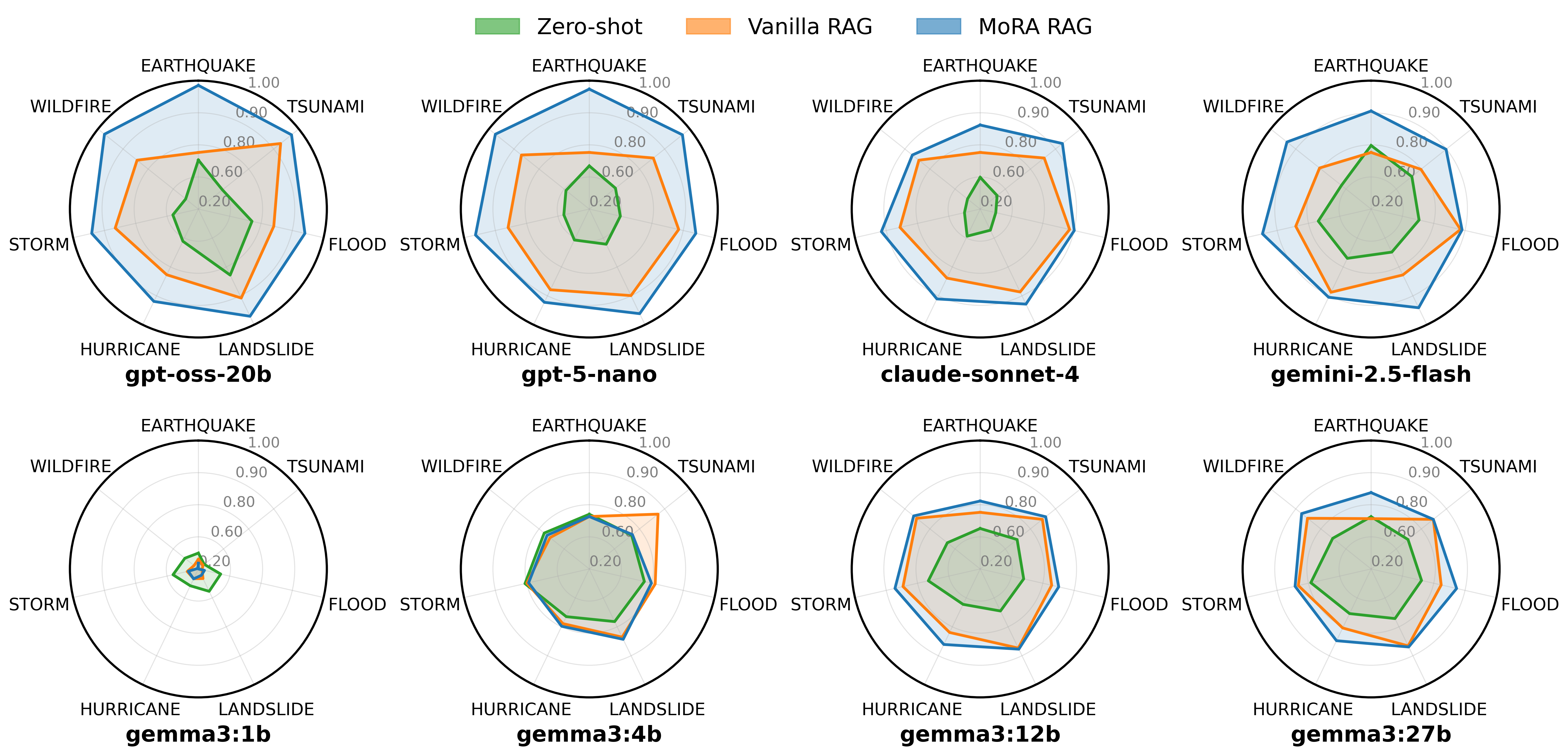}
   \caption{Model performance comparison across different backbone LLMs. First row: different models, second row: same model with different parameter sizes.}
   \label{fig:by_model}
\end{figure}

In the zero-shot setting, pretrained knowledge alone was insufficient for reliable hazard-specific understanding (Figure \ref{fig:by_model}). This limitation stems from the fact that general-purpose LLMs are not trained on detailed post-disaster observations or the mechanistic knowledge required to interpret hazard interactions, damage mechanisms, or cascading effects. The task in this study is highly specialized, requiring an understanding of multi-hazard processes and engineering context that typical pretraining corpora do not contain. Across different hazard categories, the highest accuracy reached only about 60\%, with noticeable variation among categories. The performance disparity suggests model bias, particularly for underrepresented hazards such as tsunamis and wildfires, which exhibited relatively lower accuracy.

When Vanilla RAG was introduced, model performance improved steadily across all LLMs. This gain reflects the advantage of grounding responses in domain material drawn from reconnaissance reports, which provide factual and contextual details missing from pretraining data. Nevertheless, the overall accuracy remained limited, reaching roughly 80\% at best. Although accuracy improved, some inconsistencies remained because standard retrieval approaches do not always extract complete or contextually aligned information needed for precise reasoning.

Under the proposed MoRA-RAG framework, all evaluated models showed significant improvement compared with their zero-shot counterparts. Accuracy increased to around 90\%, indicating that the framework effectively strengthened factual grounding and reasoning stability. In general, proprietary LLMs tend to outperform open-weight models due to their larger training datasets and extended optimization pipelines \citep{achiam2023gpt, kuai2025cyportqa}. However, within our experiments, the open-weight GPT-oss-20B achieved performance comparable to the proprietary GPT-5-Nano, demonstrating that well-designed retrieval and verification mechanisms can substantially reduce this performance gap. This result underscores the potential of open-weight models to serve as reliable and transparent alternatives for domain-grounded hazard reasoning. Moreover, MoRA-RAG mitigated bias across hazard categories, producing more balanced performance even for underrepresented hazards such as tsunamis and wildfires. These consistent gains across architectures of varying scale highlight the framework’s robustness and its adaptability across both open and proprietary model families.

In the second row of Figure \ref{fig:by_model}, the results show that model complexity strongly influences how effectively the framework integrates retrieved knowledge. Smaller models (i.e., Gemma3:1b) displayed irregular trends, with performance decreasing from the zero-shot setting to Vanilla RAG and further under MoRA-RAG. This pattern suggests that models with limited capacity struggle to process the expanded context and multi-agent reasoning steps, leading to information overload and reduced generation quality. Mid-sized models (i.e., Gemma3:4b) achieved partial improvement, where Vanilla RAG performed better than MoRA-RAG, implying that while retrieval provides useful context, the complete agentic workflow may exceed their optimal reasoning capacity. In contrast, larger models (Gemma3:12B and Gemma3:27B) showed a consistent improvement from the zero-shot setting to Vanilla RAG and then to MoRA-RAG. This pattern indicates that higher-capacity architectures can better utilize and integrate additional information provided by the agentic framework. Overall, these results suggest that the effectiveness of MoRA-RAG scales with model size, as larger models are more capable of handling multi-step retrieval and verification while maintaining stable reasoning performance.

In Table \ref{tab:model_perf}, we compare model performance across different question types. Under the zero-shot setting, the results show a consistent pattern across question categories, which aligns with the performance observed across hazard categories in Figure \ref{fig:by_model}. The accuracy range across the four question types, including analysis approach, hazard characteristics, impacts and damage, and response and recovery, was about 8\% across models. This range indicates that without hazard-specific knowledge, the LLMs struggle to maintain consistent reasoning across different tasks. For the RAG-based evaluation, we selected \texttt{gpt-oss-20b} as the backbone LLM because of its strong zero-shot performance and open-weight accessibility, which ensures transparency and reproducibility. All retrieval-based frameworks, including Vanilla RAG, CRAG, MAIN-RAG, and the proposed MoRA-RAG, were implemented on this backbone for consistency. Among them, MoRA-RAG achieved the highest overall accuracy of about 94.5\% and showed only around a 1\% range across question types (Table \ref{tab:model_perf}), indicating more stable and reliable performance.

\begin{table}[ht]
\centering
\footnotesize
\caption{Model performances across models and QA categories (adopted model: \texttt{gpt-oss-20b} as the agent).}
\label{tab:model_perf}
\begin{tabular*}{\linewidth}{@{\extracolsep{\fill}}l*{5}{>{\centering\arraybackslash}p{2.2cm}}}
\toprule
\textbf{Model} & \makecell{\textbf{Analysis}\\\textbf{Approach}} &
\makecell{\textbf{Hazard}\\\textbf{characteristics}} &
\makecell{\textbf{Impacts}\\\textbf{\& Damage}} &
\makecell{\textbf{Response}\\\textbf{\& Recovery}} & \textbf{Overall} \\
\midrule
\multicolumn{6}{l}{\textit{Zero-Shot}} \\
Gemma3-27B             & 72.29\% & 73.08\% & 75.78\% & 70.23\% & 73.54\% \\
gpt-oss-20b            & 68.14\% & 69.10\% & 61.57\% & 60.00\% & 65.10\% \\
gemini-2.5-flash       & 70.10\% & 74.02\% & 75.57\% & 69.32\% & 73.13\% \\
gpt-5-nano             & 63.88\% & 56.20\% & 61.54\% & 65.91\% & 60.38\% \\
claude-sonnet-4        & 51.45\% & 42.31\% & 50.00\% & 53.93\% & 49.51\% \\
\midrule
\multicolumn{6}{l}{\textit{RAG-based}~\textsuperscript{a}} \\
Vanilla RAG                  & 87.32\% & 80.65\% & \underline{83.03\%} & 88.64\% & 83.60\% \\
\underline{CRAG}\textsuperscript{b}             & \underline{88.99\%} & \underline{81.04\%} & 80.23\% & \underline{89.77\%} & \underline{85.20\%} \\
MAIN-RAG                     & 84.45\% & 79.08\% & 77.83\% & 88.64\% & 81.96\% \\
\textbf{MoRA-RAG (Proposed)}\textsuperscript{b} & \textbf{93.78\%} & \textbf{94.74\%} & \textbf{95.02\%} & \textbf{94.32\%} & \textbf{94.53\%} \\
\midrule
\multicolumn{6}{p{0.96\linewidth}}{\footnotesize a. We adopt \texttt{gpt-oss-20b} as the backbone LLM for retriever-based models due to its strong performance in open-weight models (see Figure \ref{fig:by_model}).} \\
\multicolumn{6}{p{0.96\linewidth}}{\footnotesize b. \textbf{Bold} values indicate the best-performing model per category; \underline{underlined} values indicate the second-best.} \\
\bottomrule
\end{tabular*}
\end{table}

\subsection{Ablation Studies}




\subsubsection{RQ1: Agentic Structure and Module Contribution}

The ablation study in Table \ref{tab:module_performance} shows that each agentic module contributes incrementally to performance improvement. Compared with the Vanilla RAG baseline (83.6\%), adding the mixture-of-retriever module increased accuracy by about 1\%, reflecting more targeted retrieval across hazard domains with slightly reduced latency. Incorporating the online search agent yielded a larger improvement of nearly 6\%, demonstrating the value of supplementing domain data with external knowledge when in-domain evidence is limited. The reflection and rewriter module provided an additional 8\% gain by refining ambiguous queries and strengthening context alignment. When all modules were combined, the complete MoRA-RAG framework reached an overall accuracy of 94.5\%, an improvement of almost 11\% over the baseline. Although latency increased due to the multi-step reasoning loop, the overall trade-off favors reliability and factual completeness. These results confirm that the agentic structure meaningfully improves reasoning robustness and supports the conclusion of RQ1 that coordinated agent design is critical for enhancing multi-hazard understanding.

\begin{table}[ht]
\centering
\footnotesize
\caption{Ablation study of model components}
\label{tab:module_performance}
\begin{tabular}{lcc}
\toprule
\textbf{Module Setting} & \textbf{Overall Accuracy} & \textbf{Average Latency} \\
\midrule
Vanilla RAG                   & 83.60\%                 & 4.13s  \\
RAG + MoR                     & 84.65\% (+1.05\%)       & 3.40s (-0.73s)   \\
RAG + Online Search           & 89.46\% (+5.86\%)       & 7.32s (+3.19s)   \\
RAG + Reflection \& Rewriter  & 91.23\% (+7.63\%)       & 18.56s (+14.43s)   \\
MoRA RAG (All components)     & 94.53\% (+10.93\%)      & 20.41s (+16.28s)  \\
\bottomrule
\end{tabular}
\end{table}

\subsubsection{RQ2: Chunking and Retrieval Strategy}
The comparison in Table \ref{tab:chunking_methods} highlights the strong influence of chunking strategies on retrieval accuracy. Fixed-token and paragraph-based chunking achieved comparable performance, both reaching around 91\% overall accuracy. These methods retain general context but still introduce fragmented semantics and redundant overlaps. The proposition-based method performed notably worse, with an average accuracy of about 72\%, indicating that excessive granularity weakens contextual continuity and makes retrieval more sensitive to phrasing. In contrast, the agentic chunking method achieved the highest overall accuracy of 94.5\%, improving by nearly 4\% over fixed-token and paragraph-based approaches. This method uses an LLM to combine semantically related propositions, summarize contents, and form coherent retrieval units that preserve local context while avoiding information fragmentation. The results confirm RQ2, demonstrating that adaptive, LLM-assisted chunking significantly enhances retrieval precision and semantic consistency in multi-hazard knowledge extraction.


\begin{table}[ht]
\centering
\footnotesize
\caption{Accuracy comparison of different chunking methods for RAG.}
\label{tab:chunking_methods}
\begin{tabular}{lccccc}
\toprule
\textbf{Chunking Method} & \makecell{\textbf{Analysis}\\\textbf{Approach}} & \makecell{\textbf{Hazard}\\\textbf{Characteristics}} & \makecell{\textbf{Impacts }\\\textbf{\& Damage}} & \makecell{\textbf{Response}\\\textbf{\& Recovery}} & \textbf{Overall} \\
\midrule
Fixed-token                       & 93.65\% & 89.79\% & 91.82\% & 87.95\% & 90.80\% \\
Paragraph-based                   & 91.96\% & 88.25\% & 91.14\% & 91.36\% & 90.68\% \\
Proposition-based                 & 75.77\% & 70.64\% & 71.36\% & 70.91\% & 71.92\% \\
Agentic chunk (adopted)           & \textbf{93.78\%} & \textbf{94.74\%} & \textbf{95.02\%} & \textbf{94.32\%} & \textbf{94.53\%} \\
\midrule
\multicolumn{6}{l}{\footnotesize Note: \textbf{Bold} values indicate the best-performing method per category.}\\
\bottomrule
\end{tabular}
\end{table}

\section{Conclusion}
This study introduced an agentic, knowledge-grounded LLM framework (MoRA-RAG) for improving multi-hazard understanding from reconnaissance reports. The framework was designed to address the limitations of conventional retrieval-augmented generation systems, which often suffer from fragmented retrieval, insufficient evidence grounding, and unreliable reasoning in complex hazard scenarios. Through comprehensive evaluations, MoRA-RAG demonstrated consistent performance improvements across model families, hazard categories, and question types. By integrating a mixture-of-retriever structure and an agentic verification loop, the framework enhanced retrieval precision, reasoning reliability, and factual consistency, achieving an overall accuracy of 94.5\%. The ablation study further confirmed that each agentic module (i.e., retrieval routing, online search, and reflection–rewriting) contributed to the overall gain, with coordinated operation yielding the highest performance. Additionally, results across models of varying scale revealed that open-weight LLMs such as GPT-oss-20B can reach performance levels comparable to proprietary models like GPT-5-Nano when equipped with effective grounding and verification mechanisms, narrowing a long-standing performance gap in the field \citep{achiam2023gpt}.

The findings highlight the potential of domain-grounded LLMs to transform post-disaster reconnaissance data into structured, interpretable knowledge. By enabling more reliable reasoning across heterogeneous hazard information, MoRA-RAG contributes to the growing integration of AI and disaster risk reduction research. Future work will focus on extending the framework to multi-modal datasets that combine textual, visual, and geospatial sources, allowing the model to reason jointly over imagery, maps, and sensor data~\citep{li2025leveraging,zhou2025automated}. Expanding agentic mechanisms toward adaptive learning and multi-hazards generalization also represents a promising direction for advancing data-driven resilience analysis and management.


\newpage
\appendix
\section{Details for HazardRecQA Construction}
\label{first: QA}

The HazardRecQA dataset is constructed by automatically generating QA pairs from paragraphs in reconnaissance reports. Each paragraph serves as a factual grounding source, and the generation process follows a structured system prompt designed to ensure both accuracy and exam-level clarity. Below are the implementation details:

\begin{tcolorbox}[
  colback = blue!8!white,
  colframe = blue!60!black,
  title   = {\strut QA Construction Prompt from Reconnaissance Reports},
  width   = \linewidth, boxsep=4pt, left=3pt, right=3pt, top=2pt, bottom=2pt]
\small\rmfamily
\begin{alltt}
System Prompt:
Act as a hazard expert, generate an exam-ready QA item grounded in the paragraph.
Context: The \{\textbf{{disaster_type}}\} occurred in \{\textbf{{year}}\} at \{\textbf{{location}}\}. 
A paragraph from the reconnaissance report describes the hazard: \{\textbf{{paragraph}}\}.

Rules:
1) Each question must be strictly grounded in factual evidence from the paragraph
   and reflect hazard-specific or multi-hazard knowledge.
2) The question statement must always be written as a general exam-style question,
   without mentioning the paragraph or the source text.
3) If the text is unsuitable (e.g., TOC, Acknowledgements, References):
   - Return \{ \} only.
4) Otherwise, create one QA item following the JSON format:
   - True/False → \{"\textbf{statement}": <string>, "\textbf{answer}": ”true“|”false“\}
   - MC (A–D, one correct) → \{"\textbf{question}": <string>,
                               "\textbf{options}": ["A. ...","B. ...","C. ...","D. ..."],
                               "\textbf{correct}": "A"|"B"|"C"|"D"\}
5) Categorize the question under the categories: Hazard Characteristics, Analysis
   Approach, Impacts and Damage, Response and Recovery, Invalid
6) No explanations or extra text.
\end{alltt}
\end{tcolorbox}

The generated QA items are categorized according to the type of information captured in each paragraph, ensuring multi-dimensional evaluation based on the QA dataset. The four categories, and an invalid filter are summarized below:

\begin{itemize}[noitemsep, topsep=0pt, parsep=0pt]
    \item \textbf{Hazard Characteristics}: The question describes the event itself, what happened, when and where it occurred, its magnitude, setting, or physical triggers.
    \item \textbf{Analysis Approach}: The question describes how the event was examined through data collection, instruments, surveys, models, or analytical methods.
    \item \textbf{Impacts and Damage}: The question describes the consequences or losses caused by the event, such as structural failures, economic damage, or human impacts.
    \item \textbf{Response and Recovery}: The question describes actions taken after the event, including emergency response, evacuation, repair, or long-term recovery.
    \item \textbf{Invalid}: The question is unusable, presenting vague, contextless, or referring only to figures, tables, or non-substantive content.
\end{itemize}

\section{Implementation Details for Chunking Methods}
\label{sec:chunk}

In this research, four chunking methods are compared in the ablation study: 
(1) fixed-token (2) paragraph-based (3) proposition-based and (4) agentic chunking. Each method reflects a different strategy for segmenting long and information-dense reconnaissance reports into coherent units for RAG retrieval.

The fixed-token chunking is implemented with a window size of 200 tokens, supplemented by 50 overlapping tokens before and after each segment, following standard practices in professional document processing~\citep{bhat2025rethinking}. In other words, each 200-token segment forms the core of a chunk, while the additional 50 tokens on both sides provide surrounding context for better continuity.

The paragraph-based method retains the document’s original narrative flow. Reconnaissance reports are usually organized by themes such as event overview, site observation, damage documentation, and analytical discussion; therefore, paragraph segmentation often aligns with meaningful topic boundaries.

The proposition-based chunking targets finer semantic units by utilizing an LLM for chunking. It extracts standalone factual statements from complex paragraphs, rewriting incomplete phrases into full propositions (e.g., “Observed liquefaction along quay walls” to “Investigators observed liquefaction along quay walls”). This approach captures explicit knowledge statement and produce self-contained chunks.

\begin{tcolorbox}[
  colback = gray!8!white,
  colframe = gray!60!black,
  title   = {\strut Proposition-based Chunking Prompt},
  width   = \linewidth, boxsep=4pt, left=3pt, right=3pt, top=2pt, bottom=2pt]
\small\rmfamily
\begin{alltt}
System Prompt:
Act as a hazard domain expert. Decompose the given paragraph into clear, 
self-contained propositions that can be understood independently of the source text.
Context: The \{\textbf{disaster_type}\} occurred in \{\textbf{year}\} at \{\textbf{location}\}. 
A paragraph from the reconnaissance report describes the hazard: \{\textbf{paragraph}\}.

Rules:
1) Split complex or compound sentences into minimal, simple statements.
2) Keep original wording whenever possible; ensure each statement stand alone.
3) Replace pronouns (it, they, this, that) with full entity names.
4) Preserve explicit details such as dates, times (timezone/Z), locations, agencies.
5) Output the results in JSON format as a list of propositions:
   \{"\textbf{Prop}": ["<proposition_1>", "<proposition_2>", ...]\}

Example:
Input: The Los Angeles storm occurred in 2015 at Southern California .. at 22:02Z,
  a funnel cloud was sighted near Lake Hughes and flash flooding with a car stuck 
  in rock and mudslide. .."
Output: \{"Prop": [
  "During the Los Angeles storm in 2015, at 22:02 Z on October 15, the NWS reported
  flash flooding near Lake Hughes with a car stuck in a rock and mudslide." ]\}
\end{alltt}
\end{tcolorbox}

The agentic chunking method builds aggregated representations from multiple propositions. It groups semantically related statements, such as hazard conditions, observed impacts, and contextual factors, under a concise summary that captures the main idea. This approach produces compact, context-aware chunks, resembling how experts synthesize multi-sentence observations in post-disaster analyses. 
Each chunk contains fewer than ten propositions to maintain focus and ensure internal consistency, avoiding overly long or heterogeneous segments.

This method employs two collaborating agents. The first agent performs proposition extraction, as described in the proposition-based chunking method, to identify and refine standalone factual statements. The second agent performs grouping and summarization, combining semantically related propositions and generating a short summary that encapsulates the shared context. The implementation format is illustrated below.

\begin{tcolorbox}[
  colback = gray!8!white,
  colframe = gray!60!black,
  title   = {\strut Agentic Chunk Prompt (Grouping and Summarization)},
  width   = \linewidth, boxsep=4pt, left=3pt, right=3pt, top=2pt, bottom=2pt]
\small\rmfamily
\begin{alltt}
System Prompt:
Act as a hazard domain expert. Given a list of factual propositions, group 
semantically related ones and generate concise summaries to form aggregated, 
context-aware chunks optimized for RAG.
Context: The propositions are provided as: \{\textbf{list\_of\_prop}\}.

Rules:
1) Process propositions sequentially in the given order.
2) Group consecutive propositions that are semantically related into one chunk.
   When a proposition is unrelated, finalize the chunk and start a new one.
3) Each chunk must contain no more than ten propositions.
4) For each chunk, write a short summary that captures the shared meaning.
   - Summaries must be concise and precise.
   - Preserve hazard or event details (dates, units, measurements).
   - Use consistent terminology across summaries.
5) Output the results in JSON format:
   \{
     "\textbf{Chunks}": [
       \{
         "\textbf{Summary}": "<string>",
         "\textbf{List of Propositions}": ["<prop1>", "<prop2>", ...]
       \}
     ]
   \}
6) Do not alter or invent facts; keep all proposition text unchanged.
7) Return only the JSON object, no extra explanations or commentary.
\end{alltt}
\end{tcolorbox}

For all final experiments, the agentic chunking strategy is employed as the default setting, due to its performance demonstrated in ablation studies.

\section{Implementation Details for Agents in MoRA RAG}
\label{third:agent}

This section details all agent prompts or workflow, adopted agents include MoR Router, Online Search, Evidence Evaluator, Reflection \& Question Rewriter and Answer Writer.

\begin{tcolorbox}[
  colback = blue!8!white,
  colframe = blue!60!black,
  title   = {\strut MoR Router Agent Prompt},
  width   = \linewidth, boxsep=4pt, left=3pt, right=3pt, top=2pt, bottom=2pt]
\small\rmfamily
\begin{alltt}
System Prompt:
Act as a routing expert agent. Classify a user's question into probabilities over 
natural hazard categories. These probabilities determine which hazard-specific 
RAG agent(s) should be activated.
Question: \{\textbf{Question}\}

Rules:
1) Output only a valid JSON object exactly matching the output example below.
2) Use the following hazard categories as keys.
3) Assign a normalized probability (0–1) to each category so that the total sums to 1.
4) Categories with probability >= 0.2 indicate active agents.
5) Do not include explanations, text descriptions, or additional fields.

Output Example:
\{
  "Wildfire": 0.01,
  "Storm": 0.10,
  "Landslide": 0.05,
  "Hurricane": 0.61,
  "Flood": 0.21,
  "Earthquake": 0.01,
  "Tsunami": 0.01
\}
\end{alltt}
\end{tcolorbox}

\begin{tcolorbox}[
  colback = blue!8!white,
  colframe = blue!60!black,
  title   = {\strut Online Search Agent Workflow},
  width   = \linewidth, boxsep=4pt, left=3pt, right=3pt, top=2pt, bottom=2pt]
\small\rmfamily
\begin{alltt}
Description:
The Online Search Agent performs real-time factual retrieval from external web 
sources using a online search API (i.e., DDGS). The agent is non-generative.

Workflow:
1) Receive the query and submit to the external search API (i.e., DDGS.text()).
3) Collect the top-N textual snippets (N <= 5).

Output Example:
\{
  "SearchResults": ["Harvey (2017) reached Category 4 before landfall.", ...]
\}
\end{alltt}
\end{tcolorbox}

\begin{tcolorbox}[
  colback = blue!8!white,
  colframe = blue!60!black,
  title   = {\strut Evidence Evaluator Agent Prompt},
  width   = \linewidth, boxsep=4pt, left=3pt, right=3pt, top=2pt, bottom=2pt]
\small\rmfamily
\begin{alltt}
System Prompt:
You are an expert in hazards and resilience.Decide whether the related excerpts
contain sufficient information to answer the question or evaluate the statement.
Question: \{\textbf{question}\}
Related Evidence: \{\textbf{evidence}\}

Rules:
1) Judge the factual sufficiency of the excerpts with respect to the question.
2) Do not infer or assume information beyond what is provided.
3) Output only a single value:
   - '1' if the evidence provide enough information.
   - '0' if the evidence are insufficient.
\end{alltt}
\end{tcolorbox}

\begin{tcolorbox}[
  colback = blue!8!white,
  colframe = blue!60!black,
  title   = {\strut Reflection \& Question Rewriter Agent Prompt},
  width   = \linewidth, boxsep=4pt, left=3pt, right=3pt, top=2pt, bottom=2pt]
\small\rmfamily
\begin{alltt}
System Prompt:
You are an expert in hazards and resilience with strong knowledge of information 
retrieval. Rewrite the given question to improve retrieval accuracy.
Original Question: \{\textbf{question}\}
Retrieved Insufficient Information: \{\textbf{evidence}\}

Rules:
1) Replace vague expressions with precise, domain-relevant language.
2) If the question is too broad, narrow it to highlight key entities or events.
3) If the question is too narrow, generalize slightly for broader information.
4) Maintain the original intent and semantics of the question.
5) Output only the rewritten question, no explanations or commentary.
\end{alltt}
\end{tcolorbox}

\begin{tcolorbox}[
  colback = blue!8!white,
  colframe = blue!60!black,
  title   = {\strut Answer Writer Agent Prompt},
  width   = \linewidth, boxsep=4pt, left=3pt, right=3pt, top=2pt, bottom=2pt]
\small\rmfamily
\begin{alltt}
System Prompt:
You are an expert in hazards and resilience. Determine the correct answer based
on the retrieved evidence provided by reports or online resources.
Question: \{\textbf{question}\}
Trustworthy Evidence (if applicable): \{\textbf{evidence}\}

Rules:
1) Judge strictly on the factual information in the evidence.
2) Do not infer or assume information beyond what is given.
3) Output format:
   - If the question is True/False, answer with one word: true or false.
   - If the question is Multiple Choice, answer with one letter: A, B, C, or D.
4) Do not include punctuation, explanations, or additional text.
\end{alltt}
\end{tcolorbox}

\bibliographystyle{elsarticle-num} 
\bibliography{cas-refs}

\end{document}